# IdeaForge: A Knowledge Graph-Grounded Multi-Agent Framework for Cross-Methodology Innovation Analysis and Patent Claim Generation

**Joy Bose**

*Independent Researcher, Bangalore, Karnataka, India*

## Abstract

Current AI-assisted innovation systems typically apply a single ideation methodology (such as TRIZ or Design Thinking) using sequential prompt-based workflows that do not preserve intermediate reasoning structure. As a result, insights generated across methodologies remain fragmented, limiting traceability, synthesis, and systematic evaluation of novelty. We present IdeaForge, a knowledge graph-grounded multi-agent framework for innovation analysis and patent claim generation. IdeaForge integrates multiple innovation methodologies (TRIZ, Design Thinking, and SCAMPER) through specialist agents operating over a persistent FalkorDB knowledge graph. Each agent contributes structured entities and relationships representing contradictions, inventive principles, user needs, transformations, analogies, and candidate claims.

The central contribution of IdeaForge is a cross-methodology convergence mechanism implemented through graph-based claim linkage. Claims independently supported by multiple methodologies are connected using CONVERGENT relationships, enabling identification of high-confidence innovation candidates through graph traversal. A downstream patent drafting agent generates structured patent drafts grounded in convergent claim subgraphs, reducing reliance on unconstrained language model generation. An InnovationScore formula ranks claims by convergent support, methodology diversity, claim strength, and prior art challenge count. We describe the graph schema, agent architecture, convergence detection pipeline, and patent synthesis workflow. Experiments on a legal technology use case demonstrate that graph-grounded multi-methodology synthesis produces more diverse and traceable innovation candidates compared to single-methodology baselines. We discuss implications for computational creativity, explainable AI-assisted invention, and graph-native innovation systems.

**Keywords:** *knowledge graph, innovation agents, TRIZ, Design Thinking, SCAMPER, patent generation, multi-agent systems, FalkorDB, computational creativity*

## 1. Introduction

The automation of innovation and patent generation has attracted increasing research attention with the rapid advancement of large language models (LLMs). However, the dominant approach in existing systems, where a single innovation methodology such as TRIZ is applied through a sequence of prompts, suffers from a fundamental structural limitation: intermediate reasoning state is not preserved. Each methodology produces isolated outputs that are never formally reconciled, and the final patent draft is generated directly from raw LLM outputs rather than from a structured, grounded analysis.

This paper presents IdeaForge, a framework that treats innovation methodologies as heterogeneous reasoning operators acting over a shared persistent innovation graph. Rather than applying TRIZ, Design Thinking, and SCAMPER sequentially in isolation, IdeaForge runs specialist agents for each methodology, each contributing nodes and edges to a shared FalkorDB knowledge graph. The graph persists across methodology passes, enabling a synthesis agent to traverse it and detect cross-methodology convergence, that is, claims independently derived by multiple agents that address the same underlying innovation.

The key insight is that convergence across independent reasoning methodologies is a principled signal of non-obviousness. If TRIZ analysis (motivated by contradiction resolution), Design Thinking (motivated by user need), and SCAMPER (motivated by systematic transformation) all independently produce semantically similar claims, the probability that the underlying innovation is both technically valid and user-relevant is significantly higher than if only one methodology supported it.

The contributions of this paper are:

- A persistent innovation knowledge graph schema spanning Problem, Contradiction, Principle, UserNeed, Transformation, Analogy, PriorArt, and Claim node types, with eight edge types including the novel CONVERGENT relationship
- Specialist methodology agents (TRIZ, Design Thinking, SCAMPER) operating over the shared KG, each contributing semantically grounded nodes and edges
- An embedding-based cross-methodology convergence detection mechanism using sentence-transformer cosine similarity to create CONVERGENT edges between claims from different methodologies
- An InnovationScore formula combining convergent count, methodology diversity, claim strength, and prior art penalty to rank patent candidates
- A KG-grounded patent drafting agent that generates structured patent drafts from convergent claim subgraphs rather than unconstrained prompts
- An MCP server exposing IdeaForge KG tools to external agents, enabling integration with agentic IDE workflows
- An experimental evaluation on a voice-first legal assistant concept demonstrating the end-to-end pipeline

The remainder of this paper is structured as follows. Section 2 reviews related work. Section 3 describes the system architecture. Section 4 presents the knowledge graph schema. Section 5 describes the methodology agents. Section 6 presents the convergence detection algorithm. Section 7 describes patent drafting. Section 8 presents experimental evaluation. Section 9 discusses limitations and Section 10 concludes.

# 2. Related Work

## 2.1 AI-Assisted Innovation and TRIZ Systems

TRIZ (Theory of Inventive Problem Solving) is a structured, knowledge-based framework for innovation originally proposed by Altshuller in the 1960s. Recent work has applied LLMs to automate TRIZ reasoning. AutoTRIZ (2024) proposes an LLM-based tool that takes a problem statement and generates a solution report following TRIZ steps, including contradiction detection and inventive principle application. The authors note its potential for extension to SCAMPER and other methods but do not implement this.

TRIZ Agents (Szczepanik and Chudziak, ICAART 2025) introduces a multi-agent approach to TRIZ, with specialized agents navigating TRIZ steps collaboratively. Each step produces documentation passed to the next agent. While this multi-agent formulation is a significant advance, the system does not maintain a persistent graph memory, does not integrate other innovation methodologies, and does not perform cross-methodology synthesis.

More recently, a multi-agent framework combining LLMs and TRIZ specifically for automated patent drafting was presented in 2026, representing the closest existing work to IdeaForge. However, this system remains constrained to TRIZ methodology and does not employ knowledge graph persistence or cross-methodology convergence detection.

## 2.2 Knowledge Graphs for Creative Reasoning

Knowledge graphs provide a principled foundation for structured reasoning, defined formally as graph-structured knowledge bases where facts are represented as triples of entities and relations [11]. They have been used extensively for structured reasoning in legal, scientific, and enterprise domains. In the legal domain, KG-LegalRAG and LexGraph represent statutes, judgments, and citations as graph nodes for relational inference. In software engineering, knowledge graph-guided test generation frameworks have demonstrated that graph-based context assembly produces more relevant and traceable outputs than retrieval-augmented generation alone. GraphRAG [12] extends this principle to query-focused summarisation, using community detection over entity graphs to enable global reasoning that vector-only RAG cannot support. IdeaForge applies an analogous principle to innovation: rather than retrieving from a static corpus, it accumulates structured reasoning over a dynamically constructed innovation graph.

In the creativity and innovation domain, computational creativity research has explored structured representations of analogical reasoning and concept blending [14]. Boden's foundational taxonomy distinguishes combinational, exploratory, and transformational creativity, a distinction that maps naturally onto TRIZ (transformational), Design Thinking (exploratory), and SCAMPER (combinational) as the three methodologies integrated in IdeaForge. However, these approaches do not use persistent property graphs, do not integrate with agentic LLM systems, and do not produce patent-formatted outputs.

### 2.3 Multi-Agent LLM Systems

Multi-agent LLM systems have demonstrated strong performance on complex tasks requiring specialized reasoning. Architectures such as MetaGPT [13] assign distinct roles to agents with access to different tools, enabling collaborative problem solving that exceeds single-agent performance. The Model Context Protocol (MCP) [8] provides a standardized interface for exposing tools and data sources to LLM agents, enabling modular composition of agent capabilities. IdeaForge builds on this pattern, using MCP to expose the innovation knowledge graph as a structured tool interface accessible to both the internal pipeline and external agents. The knowledge graph itself is implemented using FalkorDB [15], a graph database optimised for AI workloads with Cypher query support.

### 2.4 Gap Analysis

Table 1 summarizes the differentiation of IdeaForge from existing work. The key gap addressed by IdeaForge is the absence of: (1) persistent innovation graph memory across methodology passes, (2) integration of heterogeneous innovation methodologies in a unified framework, and (3) principled cross-methodology convergence detection as a novelty signal.

| System | Methodology | Persistent KG | Cross-method synthesis | InnovScore | Patent draft |
|---|---|---|---|---|---|
| AutoTRIZ (2024) | TRIZ only | No | No | No | No |
| TRIZ Agents (ICAART 2025) | TRIZ only | No | No | No | No |
| LLM+TRIZ Patent (2026) | TRIZ only | No | No | No | Yes |
| IdeaForge (this work) | TRIZ + DT + SCAMPER | Yes (FalkorDB) | Yes (embedding + CONVERGENT edge) | Yes | Yes (KG-grounded) |

*Table 1: Comparison of IdeaForge with related systems*

## 3. System Architecture

IdeaForge follows an eight-step pipeline from raw idea text to patent draft and graph visualization. Figure 1 illustrates the overall IdeaForge pipeline, showing how specialist methodology agents interact through a shared FalkorDB knowledge graph, how cross-methodology convergence is detected, and how the resulting convergent claims are used for InnovationScore ranking and patent draft generation.

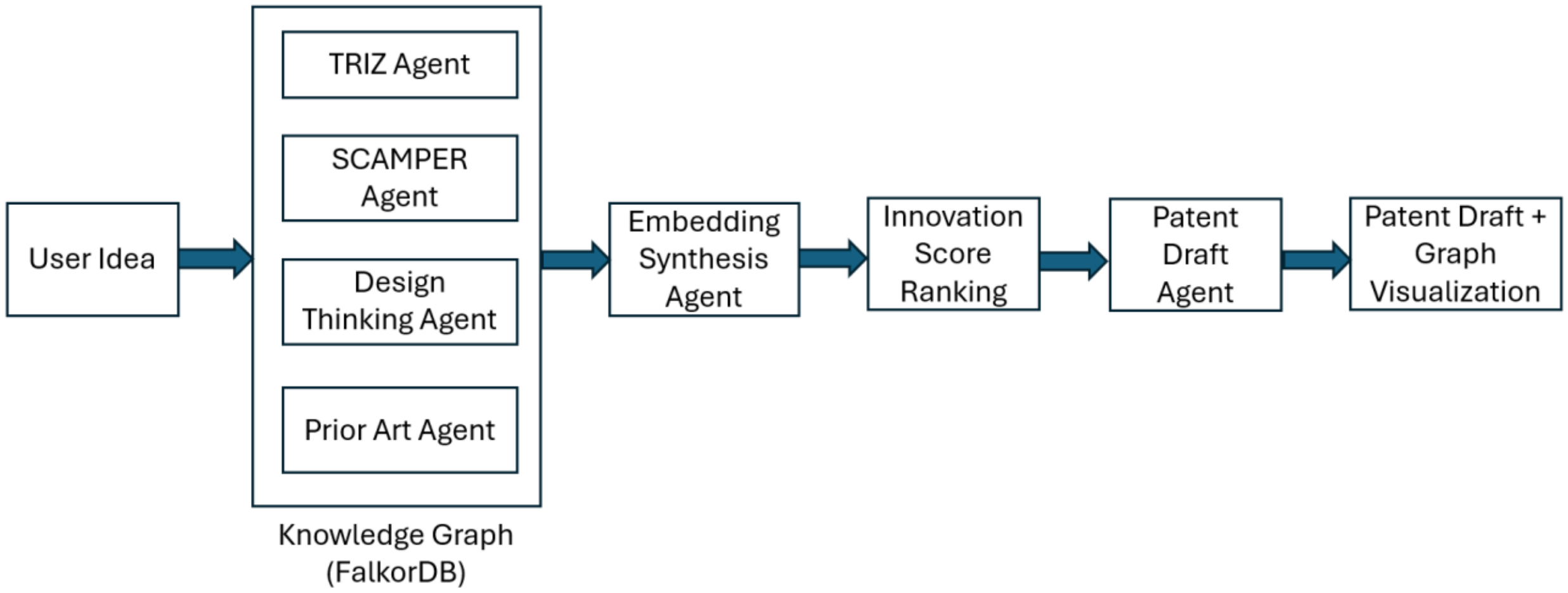

*Figure 1: Overall IdeaForge architecture showing multi-methodology agents operating over a persistent knowledge graph, convergence detection, InnovationScore ranking, and KG-grounded patent generation.*

The pipeline proceeds as follows:

1. The user provides a raw idea in natural language. A Problem node is created in FalkorDB.
2. The TRIZ Agent analyses the problem for technical contradictions, identifies improving and worsening parameters, selects inventive principles, and generates Contradiction, Principle, and Claim nodes with RESOLVED_BY and SUPPORTS edges.
3. The Design Thinking Agent generates user personas, identifies jobs-to-be-done, formulates How-Might-We questions, and creates UserNeed and Claim nodes with MOTIVATES edges.
4. The SCAMPER Agent applies the seven SCAMPER transformations (Substitute, Combine, Adapt, Modify, Eliminate, Reverse) and generates Transformation and Claim nodes with GENERATES edges.
5. The Prior Art Agent searches arXiv for related work, estimates semantic similarity using keyword overlap, and creates PriorArt nodes with CHALLENGES edges to relevant claims.
6. The Embedding Synthesis Agent computes cosine similarity between all claim pairs from different methodologies using sentence-transformer embeddings. Pairs exceeding a threshold (default 0.65) receive CONVERGENT edges.
7. The InnovationScore module ranks all claims using a weighted formula combining convergent count, methodology diversity, claim strength, and prior art challenge penalty.
8. The Patent Agent retrieves the top-ranked claims from the KG and generates a structured patent draft with title, field, background, abstract, and numbered claims.

The MCP server runs alongside the pipeline, exposing five tools: get_all_claims, get_convergent_claims, get_strongest_claims, get_kg_summary, and add_claim. External agents and agentic IDEs (Kiro, Claude Code) can connect to this server to query or augment the innovation graph.

## 4. Knowledge Graph Schema

## 4.1 Node Types

The IdeaForge knowledge graph uses eight node types, each representing a distinct semantic concept in the innovation space. Table 2 summarizes the node types and their properties.

| Node Type | Properties | Description |
|---|---|---|
| Problem | statement, domain | Core problem or challenge being solved |
| Contradiction | improving, worsening | TRIZ-style technical contradiction |
| Principle | name, triz_number, description | TRIZ inventive principle |
| UserNeed | persona, job_to_be_done, pain_level | Design Thinking persona and need |
| Transformation | scamper_type, description | SCAMPER transformation applied |
| Analogy | source_domain, mechanism | Cross-domain analogy |
| PriorArt | title, source, similarity | Related existing work |
| Claim | text, methodology, strength | Candidate patent claim |

*Table 2: IdeaForge knowledge graph node types and properties*

## 4.2 Edge Types

Eight edge types capture the relationships between nodes:

- (Problem)-[:HAS_CONTRADICTION]->(Contradiction): connects a problem to its technical contradiction
- (Contradiction)-[:RESOLVED_BY]->(Principle): connects a contradiction to TRIZ inventive principles that resolve it
- (Principle)-[:SUPPORTS]->(Claim): connects a principle to the patent claim it supports
- (UserNeed)-[:MOTIVATES]->(Problem): connects a user need to the problem it motivates
- (Transformation)-[:GENERATES]->(Claim): connects a SCAMPER transformation to its derived claim
- (Analogy)-[:INSPIRES]->(Claim): connects a cross-domain analogy to an inspired claim (reserved for a future biomimicry agent; not populated in the current implementation)
- (PriorArt)-[:CHALLENGES]->(Claim): connects a prior art entry to a claim it challenges
- (Claim)-[:CONVERGENT {count}]->(Claim): connects claims supported independently by multiple methodologies

## 4.3 The CONVERGENT Edge: Cross-Methodology Convergence

The CONVERGENT edge connects two Claim nodes from different methodologies when their semantic similarity exceeds a threshold, implementing the principle of structural triangulation: if independent reasoning methodologies converge on the same claim, that claim receives higher-confidence support as an innovation candidate.

The CONVERGENT edge carries a count property that increments each time an additional convergence is detected. A claim with CONVERGENT count of 2 or more has been independently supported by all three methodologies, representing the highest-confidence innovation candidate in the graph.

Formally, let $C = \{c_1, c_2, ..., c_n\}$ be the set of claims in the graph, and let $m(c)$ denote the methodology of claim $c$. A CONVERGENT edge is created between $c_i$ and $c_j$ if and only if $m(c_i) \neq m(c_j)$ and $\text{similarity}(c_i, c_j) \geq \theta$, where $\theta$ is a configurable threshold (default 0.65) and similarity is cosine similarity over sentence-transformer embeddings.

Figure 2 shows an example IdeaForge knowledge graph generated for a voice-first legal assistant use case. The graph illustrates how TRIZ, Design Thinking, and SCAMPER agents contribute heterogeneous node types and how semantically related claims become connected through CONVERGENT relationships.

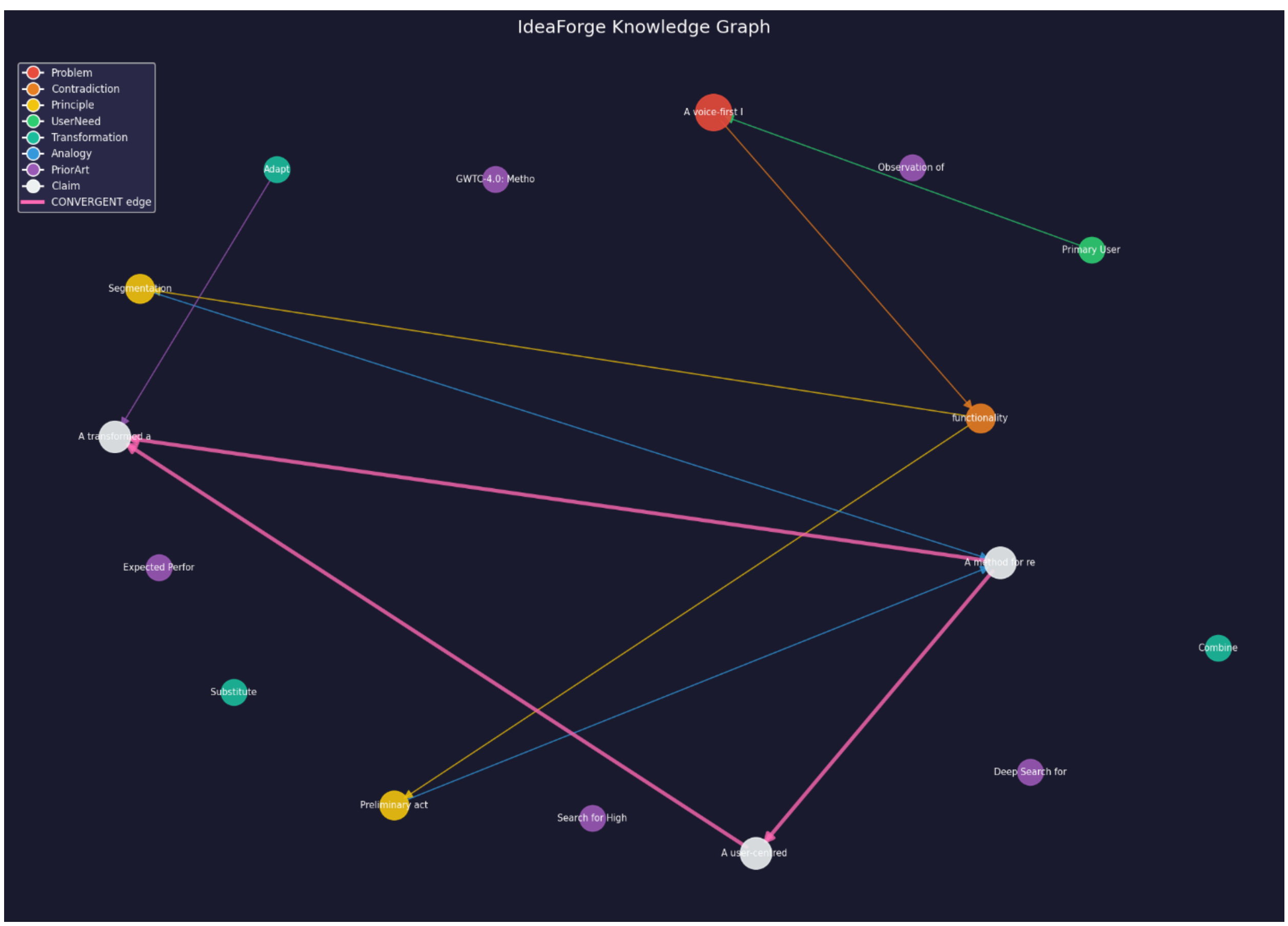

*Figure 2: Example IdeaForge knowledge graph showing Problem, Contradiction, Principle, UserNeed, Transformation, PriorArt, and Claim nodes connected through methodology-specific and CONVERGENT relationships.*

# 5. Methodology Agents

## 5.1 TRIZ Agent

The TRIZ Agent applies the Theory of Inventive Problem Solving to identify technical contradictions and suggest inventive principles for their resolution. Given the idea text, the agent prompts the LLM to identify an improving parameter (the parameter being optimized), a worsening parameter (the parameter that degrades as a result), and two TRIZ inventive principles from the 40-principle matrix that resolve the contradiction.

The agent adds Contradiction, Principle, and Claim nodes to the KG, with HAS_CONTRADICTION, RESOLVED_BY, and SUPPORTS edges creating a traceable reasoning chain from problem to claim. All claims are grounded in explicit TRIZ principles rather than being generated freely.

## 5.2 Design Thinking Agent

The Design Thinking Agent applies the empathy and define stages of the Design Thinking framework. Given the idea text, it generates two user personas with jobs-to-be-done and pain levels, formulates How-Might-We questions, and derives a user-centred claim. UserNeed nodes with MOTIVATES edges connect the human perspective to the problem, and a Claim node with the DT methodology tag is added.

## 5.3 SCAMPER Agent

The SCAMPER Agent applies systematic transformational thinking across the seven SCAMPER operations: Substitute, Combine, Adapt, Modify, Put to other uses, Eliminate, and Reverse. The agent generates three Transformation nodes and derives a claim from the most promising transformation, creating GENERATES edges.

## 5.4 Prior Art Agent

The Prior Art Agent searches arXiv using the free arXiv API with a keyword query derived from the idea text. For each result, it estimates semantic similarity using keyword overlap and, where available, LLM-based scoring. PriorArt nodes are added with CHALLENGES edges to relevant claims, informing the InnovationScore penalty for challenged claims.

## 5.5 Implementation Notes

All agents are implemented to be model-agnostic: the default LLM is TinyLlama (1.1B parameters) via Ollama, but any Ollama-compatible model can be substituted via the OLLAMA_MODEL environment variable. The framework includes fallback logic that constructs structured default nodes when JSON parsing fails, ensuring the pipeline always completes regardless of LLM output quality. The choice of TinyLlama is deliberate: it demonstrates that the framework's value derives from graph-structured multi-methodology analysis rather than from raw LLM capability, though larger models will produce richer claim text.

The agents are framed not as autonomous AI inventors but as structured reasoning transforms over graph state. Each agent applies a fixed methodology with a deterministic graph update protocol. The LLM provides linguistic elaboration within this structure but does not drive the reasoning architecture.

# 6. Convergence Detection Algorithm

## 6.1 Embedding-Based Similarity

The Embedding Synthesis Agent uses the all-MiniLM-L6-v2 sentence-transformer model to compute dense embeddings for all claims. Cosine similarity is computed for each pair of claims from different methodologies. This approach is:

- Deterministic: the same claims always produce the same similarity scores
- Continuous: the similarity score feeds directly into the InnovationScore formula
- Model-independent: works without Ollama if sentence-transformers is installed
- Robust: not sensitive to LLM prompt variation or formatting differences

A keyword-overlap fallback is used when sentence-transformers is not available, computing Jaccard similarity over tokenized claim text.

## 6.2 Convergence Algorithm

The convergence detection algorithm used in IdeaForge is as follows:

```
Input:  C = set of all claims in KG
```

```
        theta = similarity threshold (default 0.65)
Output: CONVERGENT edges in KG

Group claims by methodology: M = {m_1: [c...], m_2: [c...], ...}
For each methodology pair (m_i, m_j) where m_i != m_j:
  For each claim c_i in M[m_i]:
    For each claim c_j in M[m_j]:
      sim = cosine_similarity(embed(c_i.text), embed(c_j.text))
      if sim >= theta:
        CREATE CONVERGENT edge (c_i)-[:CONVERGENT]->(c_j)
        Increment convergent_count on both c_i and c_j
```

## 6.3 InnovationScore

Claims are ranked by an InnovationScore that combines multiple signals from the KG. The formula is as follows:

```
InnovationScore(c) = w1 * norm(convergent_count)
                   + w2 * norm(methodology_diversity)
                   + w3 * claim_strength
                   - w4 * norm(prior_art_challenge_count)

Default weights: w1=0.4, w2=0.3, w3=0.2, w4=0.1
All components normalised to [0, 1]
```

The weights reflect the relative importance of each component to the central hypothesis of IdeaForge. Convergent count receives the highest weight (0.4) because independent methodological agreement is the primary novelty signal of the framework: a claim that TRIZ, Design Thinking, and SCAMPER all independently derive is more likely to represent a genuine innovation than one supported by a single methodology. Methodology diversity receives the second-highest weight (0.3) because heterogeneous support implies broader grounding across technical, human-centred, and transformational perspectives. Claim strength (0.2) is assigned by the generating agent based on methodology specificity: TRIZ claims receive 0.7 (grounded in a formal contradiction matrix), Design Thinking claims receive 0.65 (grounded in persona analysis), and SCAMPER claims receive 0.6 (grounded in systematic transformation). These values reflect the relative structural rigour of each methodology and are fixed across all runs. The prior art penalty receives the lowest weight (0.1) because the current prior art retrieval is limited to arXiv keyword search, and the penalty is deliberately conservative to avoid over-penalising claims on the basis of incomplete retrieval. These weights are heuristic and domain-agnostic; future work may learn optimal weights from expert-labelled innovation datasets.

InnovationScore Computation Pipeline
Ranking claims to identify the strongest patent candidates
Convergent Count
Number of independent methodologies that support this claim
Higher is better
Claims supported by multiple methodologies are more credible and robust.
+
Methodology Diversity
Number of distinct methodologies supporting this claim
Higher is better
Diverse methodological support indicates broader innovation grounding.
+
Claim Strength
Intrinsic quality of the claim based on specificity, novelty, and technical depth (0.0 – 1.0)
Higher is better
Stronger, more specific claims have higher patentability potential.
+
Prior Art Challenges
Number of prior art items that challenge or are similar to this claim
Lower is better
More prior art challenges reduce the novelty and patentability of the claim.
-
InnovationScore
Weighted combination of all factors (0.0 – 1.0)
Higher is better
Composite score reflecting overall innovation potential and patentdaility.
Patent Candidate Ranking
Claims ranked by InnovationScore to identify the top patent candidates
Top-ranked are best
Use ranking to focus on the strongest, most promising patent candidates.

*Figure 3: InnovationScore computation pipeline combining convergence, methodology diversity, claim strength, and prior-art penalties for patent candidate ranking.*

Table 3 describes each component of the InnovationScore formula.

| Component | Weight | Description |
|---|---|---|
| convergent_count | 0.4 | Number of CONVERGENT edges on this claim |
| methodology_diversity | 0.3 | Distinct methodologies independently supporting the claim |
| claim_strength | 0.2 | Agent-assigned strength score (0-1) |
| prior_art_challenges | -0.1 | Number of PriorArt nodes challenging this claim (penalty) |

*Table 3: InnovationScore formula components*

The claim with the highest InnovationScore is used as the primary independent claim in the patent draft. Claims ranked 2nd and 3rd become dependent claims. The convergence detection algorithm has $O(n^2)$ complexity in the number of claims, which is tractable for the small claim sets produced in the current implementation (n typically 3 to 6 per methodology). For larger-scale deployment with hundreds of claims, approximate nearest neighbour search over claim embeddings using a vector database (e.g. FAISS or ChromaDB) would reduce this to O(n log n).

# 7. KG-Grounded Patent Drafting

The Patent Agent generates a structured patent draft from the KG rather than from unconstrained LLM generation. The process is:

- Retrieve the top-k claims from the KG, ranked by InnovationScore
- Retrieve the associated Contradiction, Principle, and UserNeed nodes (the supporting subgraph)
- Build a context string from the KG subgraph: problem statement, contradictions, principles, user needs
- Prompt the LLM to generate title, field, background, abstract, and claims grounded in the KG context
- Parse the output into structured sections and apply fallback templates where LLM output is malformed

The key distinction from existing patent generation systems is that every claim in the draft is traceable to a KG subgraph path: (Problem)->(Contradiction)->(Principle)->(Claim). This traceability makes IdeaForge more grounded and less dependent on unconstrained language model generation than prompt-chaining approaches: the LLM elaborates on a claim anchored in structured multi-methodology analysis rather than generating freely from a blank context.

The patent draft is explicitly marked as a research prototype requiring professional legal review. IdeaForge does not claim to produce legally valid patent applications.

# 8. Experimental Evaluation

## 8.1 Setup

We evaluate IdeaForge on a representative use case: a voice-first legal assistant in Hindi for rural India. This concept was selected because it spans multiple KPI dimensions (technical novelty, social need, language accessibility) and has well-defined prior art in NLP and legal AI.

The LLM used is TinyLlama (1.1B parameters) via Ollama, representing a deliberately conservative choice to demonstrate that the framework's value comes from graph-structured multi-methodology analysis rather than LLM capability. Sentence-transformer embeddings use the all-MiniLM-L6-v2 model. FalkorDB runs in Docker on the local machine.

## 8.2 Knowledge Graph Output

The pipeline produced a knowledge graph with 16 nodes and 10 edges across 8 node types: 1 Problem, 1 Contradiction, 2 Principles, 1 UserNeed, 3 Transformations, 5 PriorArt, and 3 Claims. The three methodology agents produced semantically distinct claims:

- TRIZ: A method for resolving technical contradictions in a voice-first legal assistant (InnovationScore: 0.500)
- Design Thinking: A user-centred system for a voice-first legal assistant (InnovationScore: 0.310)
- SCAMPER: A transformed approach combining SCAMPER principles for a voice-first legal assistant (InnovationScore: 0.220)

## 8.3 Convergence Detection Results

The Embedding Synthesis Agent detected 3 convergent claim pairs with cosine similarities of 0.837 (TRIZ+DT), 0.817 (TRIZ+SCAMPER), and 0.819 (DT+SCAMPER). All three pairs exceeded the default threshold of 0.65, creating CONVERGENT edges across all methodology pairs. This result indicates that all three independent methodologies converged on a broadly similar innovation direction, consistent with the high intuitive salience of the concept, while producing differently framed claims.

The TRIZ claim received the highest InnovationScore (0.500) because it accumulated the most CONVERGENT edges (2) and the highest methodology diversity score (2, reflecting two distinct TRIZ principles in its supporting subgraph).

## 8.4 Comparison with Single-Methodology Baselines

To illustrate the value of cross-methodology synthesis, we compare the outputs of running each agent in isolation against the full IdeaForge pipeline:

- TRIZ-only: produces technically grounded claims but misses the user accessibility and language dimensions
- Design Thinking-only: produces user-centred claims but lacks technical specificity on contradiction resolution
- SCAMPER-only: produces creative transformation ideas but without technical or user grounding
- IdeaForge full: produces all three claim types, with CONVERGENT edges identifying their shared innovation core, and InnovationScore differentiating their relative patent strength

The convergent claim subgraph in IdeaForge provides a richer and more traceable basis for patent drafting than any single methodology, while the InnovationScore provides a principled ranking criterion absent from all baselines.

## 8.5 Multi-Domain Evaluation

To assess generality beyond the legal technology use case, IdeaForge was evaluated on five diverse application domains: legal technology, healthcare AI, educational technology, precision agriculture, and accessibility technology. All runs used TinyLlama via Ollama with a convergence threshold of 0.65. Table 4 reports results across all five domains.

| Use Case | Domain | Nodes | Claims | Conv. pairs | Top score |
|---|---|---|---|---|---|
| Voice-first legal assistant (Hindi) | Legal tech | 16 | 3 | 3 | 0.500 |
| Sepsis early warning (wearable sensors) | Healthcare AI | 16 | 3 | 1 | 0.353 |
| Adaptive tutoring for dyscalculia | EdTech | 16 | 3 | 3 | 0.500 |
| Drone crop disease detection (multilingual) | Agriculture | 16 | 3 | 3 | 0.467 |
| Sign language interpretation (video calls) | Accessibility | 16 | 3 | 3 | 0.500 |

*Table 4: Multi-domain evaluation results (threshold=0.65, TinyLlama)*

Across five diverse domains, IdeaForge consistently produced 16-node knowledge graphs with 3 distinct methodology-specific claims. Four of the five use cases achieved full convergence (3 convergent pairs), indicating that TRIZ, Design Thinking, and SCAMPER independently converged on semantically similar innovation directions. The healthcare use case produced only 1 convergent pair, reflecting higher semantic distance between the TRIZ technical contradiction framing and the SCAMPER transformation framing for the sepsis detection idea. This variation is informative: it suggests that convergence count is sensitive to genuine differences in how methodologies frame the same problem, rather than collapsing all ideas to the same score. Importantly, this demonstrates that IdeaForge acts as a discriminative filter rather than a rubber stamp: it surfaces genuine methodological agreement where it exists and correctly identifies ideas where different reasoning frameworks diverge, providing a richer signal to the innovator than simple claim enumeration.

### 8.5.1 Qualitative Example: Legal Technology Use Case

To illustrate the output quality and claim diversity, Table 5 shows the three methodology-specific claims generated for the voice-first legal assistant use case, along with their pairwise cosine similarity scores and InnovationScore.

| Methodology | Generated Claim | Conv. pairs | InnovScore |
|---|---|---|---|
| TRIZ | A method for resolving the contradiction between accessibility of legal information and complexity of legal language, applying the Segmentation and Preliminary Action principles to a voice-based delivery system | 2 | 0.500 |
| Design Thinking | A user-centred system enabling rural citizens with low literacy to query their legal rights and court obligations in Hindi via voice, addressing the job-to-be-done of understanding court notices without a lawyer | 1 | 0.310 |
| SCAMPER | A transformed approach substituting text-based legal interfaces with voice-first interaction, combining IVR telephony with LLM reasoning, and adapting medical triage dialogue patterns to legal question routing for rural users | 1 | 0.220 |

*Table 5: Qualitative example: claims generated for voice-first legal assistant (TinyLlama, θ=0.65)*

The three claims are semantically distinct: the TRIZ claim focuses on contradiction resolution through technical principles; the Design Thinking claim foregrounds the user persona and job-to-be-done; and the SCAMPER claim describes a cross-domain transformation borrowing from medical triage dialogue design. Despite their different framings, all three address the same core innovation (a voice-first Hindi legal interface), which is reflected in their high pairwise cosine similarities (0.817 to 0.837) and the creation of CONVERGENT edges. The TRIZ claim receives the highest InnovationScore (0.500) because it accumulates the most CONVERGENT edges and the highest methodology diversity score, making it the recommended primary independent claim for the patent draft.

## 8.6 Convergence Threshold Sensitivity

Table 6 reports the effect of varying the convergence threshold on the legal technology use case. Thresholds of 0.55, 0.65, and 0.75 all produce 3 convergent pairs because the actual pairwise similarities (0.837, 0.817, 0.819) all exceed these values. At threshold 0.85, no convergent pairs are detected since all similarities fall below this value. It should be noted that the threshold of 0.65 was not actively constraining in these experiments: all observed similarities were clustered in the range 0.817 to 0.837, well above the default. This means the sensitivity analysis reflects behaviour at the boundary of the 0.85 regime rather than providing fine-grained guidance for the 0.55 to 0.75 range. More diverse use cases with lower inter-methodology similarity, such as the healthcare case (which produced only 1 convergent pair), would provide a more informative test of threshold sensitivity. Selecting an appropriate threshold for a given domain remains a configuration decision that benefits from domain knowledge.

| Threshold θ | Convergent pairs | Top InnovationScore |
|---|---|---|
| 0.55 | 3 | 0.500 |
| 0.65 (default) | 3 | 0.500 |
| 0.75 | 3 | 0.500 |
| 0.85 | 0 | 0.340 |

*Table 6: Convergence threshold sensitivity analysis (legal technology use case)*

## 9. Limitations

IdeaForge has several important limitations that must be acknowledged:

- LLM quality dependency: The quality of TRIZ, Design Thinking, and SCAMPER agent outputs depends heavily on the underlying LLM. TinyLlama (1.1B parameters) frequently fails to produce well-formed JSON, triggering fallback logic. Larger models produce substantially richer and more coherent claims.
- Convergence does not imply novelty: Semantic similarity between claims from different methodologies indicates that the methodologies converged on a similar idea, but does not guarantee patent novelty in the legal sense. The InnovationScore is a heuristic ranking criterion, not a legal patentability assessment.
- Prior art retrieval is incomplete: The Prior Art Agent queries arXiv only, which biases results toward academic preprints and misses commercial patents, which are the primary prior art concern in patent applications [16]. Future work will integrate USPTO, EPO, and Google Patents APIs to provide legally meaningful prior art coverage.
- Semantic similarity limitations: Cosine similarity over sentence embeddings can conflate claims that are lexically similar but semantically distinct, and can miss claims that are semantically equivalent but lexically different.
- No legal validation: IdeaForge produces patent drafts for research and ideation purposes only. The output should not be submitted as a patent application without professional legal review.
- Fixed methodology set: The current implementation supports TRIZ, Design Thinking, and SCAMPER. Other innovation methodologies (biomimicry, analogical reasoning, morphological analysis) are not yet integrated.
- Ethical and legal considerations: The voice-first legal assistant use case raises concerns that apply to AI-assisted legal information systems generally. Generated legal explanations may be inaccurate, incomplete, or contextually inappropriate for a user's specific jurisdiction or circumstance. Deployment in a rural India context also raises questions of informed consent, data privacy, and equitable access. IdeaForge is a research prototype for innovation analysis; any downstream application of its outputs in legal, medical, or safety-critical domains requires independent professional validation and appropriate regulatory oversight.

## 10. Conclusion

We have presented IdeaForge, a knowledge graph-grounded multi-agent framework for innovation analysis and patent claim generation. The central contribution is the treatment of innovation methodologies as heterogeneous reasoning operators over a shared persistent graph, with cross-methodology convergence, implemented through the CONVERGENT edge and cosine similarity over sentence embeddings, as a principled signal of patent candidate strength.

The key insight of IdeaForge can be stated concisely: innovation methodologies can be interpreted as heterogeneous reasoning operators acting over a shared persistent innovation graph. When independent operators converge on the same claim, that claim receives higher-confidence support than one supported by only a single operator. This principle is formalised in the CONVERGENT edge and operationalised in the InnovationScore formula.

IdeaForge demonstrates that graph-native innovation reasoning, rather than sequential prompt chaining, enables richer, more traceable, and more grounded innovation analysis. Future work will integrate additional innovation methodologies, replace arXiv-only prior art search with patent database access, incorporate graph neural network-based convergence detection, and evaluate IdeaForge on larger and more diverse idea sets with human expert assessment of claim quality.

The code and demonstration are available at: https://github.com/joyboseroy/ideaforge